\documentclass[letterpaper, 10 pt, conference]{ieeeconf}

\IEEEoverridecommandlockouts
\usepackage{iftex}
\ifPDFTeX
  \usepackage[T1]{fontenc}
\else
  \usepackage{fontspec}
\fi
\usepackage{cite}
\usepackage{url}
\usepackage[top=25.4mm, bottom=19.1mm, left=19.1mm, right=19.1mm]{geometry}
\usepackage{amsmath,amssymb,amsfonts}
\usepackage{graphicx}
\usepackage{textcomp}
\usepackage{xcolor}
\usepackage{comment}
\usepackage{balance}
\usepackage{subcaption}
\usepackage{tabularx}
\usepackage{booktabs}
\usepackage[]{siunitx}
\usepackage[hidelinks]{hyperref}
\def\BibTeX{{\rm B\kern-.05em{\sc i\kern-.025em b}\kern-.08em
    T\kern-.1667em\lower.7ex\hbox{E}\kern-.125emX}}

\begin{document}
\bstctlcite{BSTcontrol}

\title{
    Language-Grounded Force/Torque Pretraining \\ for Extrinsic Contact Perception
    \thanks{$^1$ The University of Tokyo}
    \thanks{Correspondence to: kazuki.takahashi@akg.t.u-tokyo.ac.jp}
}
\author{
    Kazuki Takahashi$^1$, Hiroaki Murakami$^1$, Ryota Sakuma$^1$, \\
    Mitsuhiro Kamezaki$^1$, Matthew Ishige$^1$, Yoshihiro Kawahara$^1$
}

\maketitle

\begin{abstract}
Fine manipulation requires robots to recognize when a grasped object makes or changes contact with its environment, yet such extrinsic contact is not always directly observable from vision or local contact geometry. 
Force and torque signals carry this information, but also vary with object weight, grasp configuration, and robot motion, leaving it unclear whether they can provide a reusable representation of physical state. 
Here we show that bilateral six-axis force/torque histories can be pretrained as a semantically structured physical representation by aligning them with a frozen language space using supervision generated from simulated pick-and-place interactions. 
On unseen objects in simulation, the learned representation supports text-prototype readout of robot action, object weight, and grasp interaction. 
In real-world placement, unlabeled calibration enables contact and support-state recognition without updating the encoder, outperforming a vertical-force sign-reversal rule baseline. 
For connector insertion, an event absent from pretraining, low-rank adaptation improves data-efficient learning over training from scratch. 
These results show that force/torque histories can support both direct readout of known physical states and efficient adaptation to new extrinsic-contact events.

\end{abstract}

\begin{figure*}[t]
    \centering
    \includegraphics[width=\linewidth]{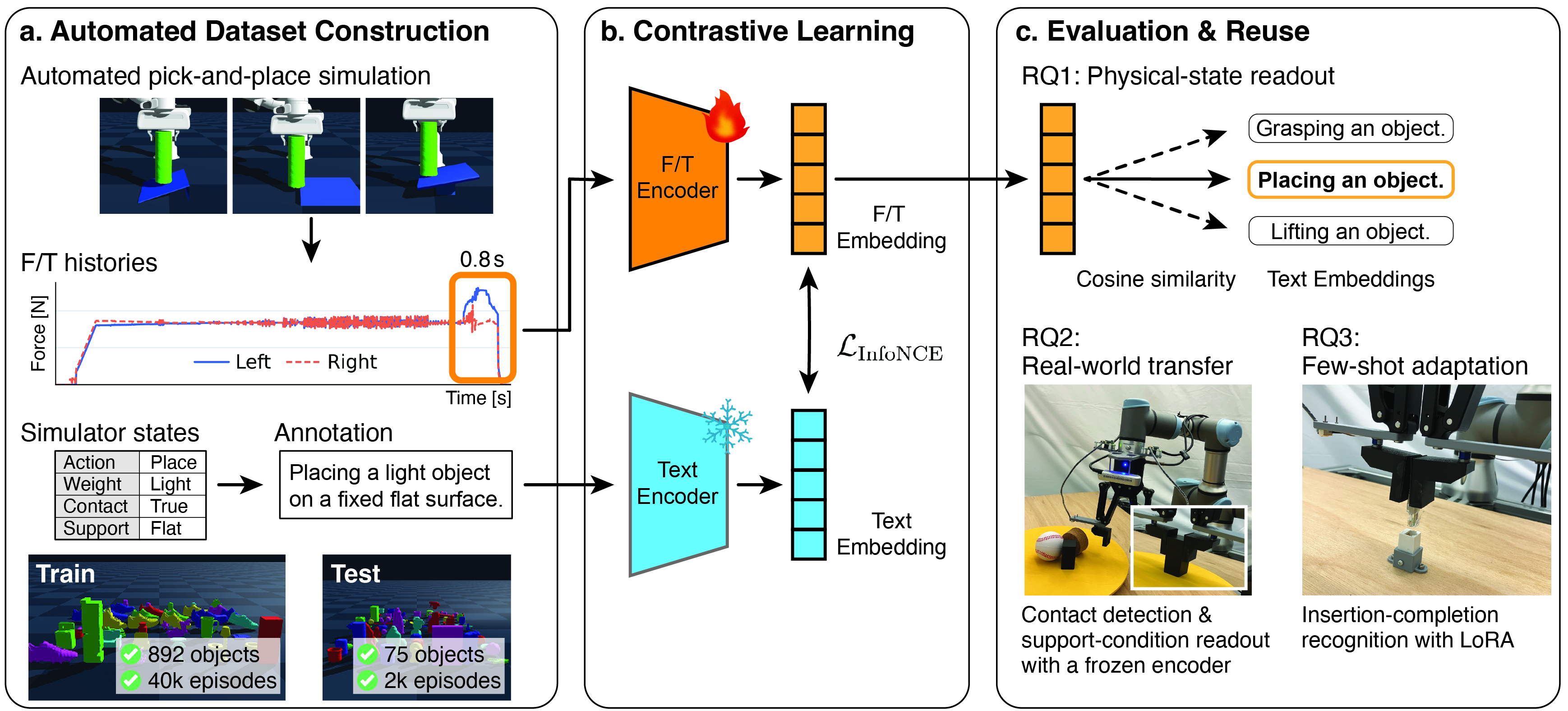}
    \caption{
    Overview. (a) Automated pick-and-place simulation generates bilateral F/T histories and semantic annotations from simulator states. (b) F/T embeddings are contrastively aligned with a frozen language space. (c) The learned representation is evaluated through physical-state readout on unseen objects, real-world transfer with a frozen encoder, and few-shot adaptation to an unseen insertion-completion event with LoRA.
    }
    \label{fig:overview}
\end{figure*}

\section{Introduction}
\label{sec:introduction}
When should a robot stop descending while placing an object on a table? How can it tell when a connector is fully seated?
Both require recognizing changes in \emph{extrinsic contact}: contact between a grasped object and the environment away from the fingertips.
Reliable perception of such contacts is critical for fine manipulation, yet their effects are not always directly observable from vision or local contact geometry.
They are transmitted as force and torque (F/T) through the grasped object, but the resulting signals also vary with object weight, grasp configuration, and robot motion.
This raises a central question: Can a force representation capture the underlying physical state well enough to be reused across objects, conditions, and contact events?

Here, we address this question through pretraining on bilateral F/T histories.
Each sensor measures a resultant force and torque combining gravity, inertia, grasping, and environmental contact.
We seek a shared representation from which action, load, grasp interaction, and support conditions can be read out. 
The input consists solely of low-dimensional F/T histories, without contact-surface images or point clouds. 
The intended reuse includes recognizing known states under new conditions and learning events absent from pretraining.

Unlike our approach, semantic contact representations have been studied using optical tactile images and 3D point clouds.
CLTP and FG-CLTP align tactile 3D representations with contact descriptions~\cite{CLTP, FG-CLTP}, while Tejwani et al.\ align human-applied force with language~\cite{ForceLanguage} and FM-VLA pretrains F/T histories for downstream action generation~\cite{FM-VLA}.
These studies establish the value of semantic or pretrained contact representations, but the physical semantics and reusability of six-axis F/T histories themselves remain unclear.
Our focus is the independent evaluation of F/T histories as a physical-state representation: whether they support multiple semantic readouts, transfer to real extrinsic contact, and adaptation to an unseen event.

We ground F/T representations in a pretrained language space to make them semantically structured and directly queryable.
Language provides both semantic supervision during pretraining and a means to specify physical states at inference.
We contrastively align F/T encoder outputs with a frozen language space, allowing defined physical states to be read out through text-prototype similarity without separate classification heads.
Our contribution concerns the physical information and reuse supported by this formulation, rather than a new encoder architecture or loss.

A major obstacle to scaling this semantic learning is the lack of semantic supervision for F/T histories.
Collecting real interactions across diverse objects and temporally labeling contact and slip intervals is costly.
We therefore use simulations to generate both F/T observations and window-level semantic descriptions from privileged physical states, enabling large-scale F/T--language pretraining without manual sequence annotation.

In evaluation, we first probe robot action, object weight, and grasp-interaction recognition on unseen objects, comparing encoder architectures and pretraining objectives.
We then evaluate two forms of reuse.
In real-object placement, the simulation-pretrained encoder is transferred to real F/T measurements after unlabeled calibration, enabling contact and support-state readout without encoder updates.
For full connector seating, a new downstream event not represented in pretraining, we use separately collected labeled data and find that frozen linear readout offers limited benefit, whereas LoRA~\cite{LoRA} adaptation improves low-data performance.

Our contributions are:
\begin{itemize}
\item We demonstrate the semantic utility of bilateral F/T histories: a common language-aligned representation supports multiple physical-state readouts through text prototypes.
\item We build an automated pipeline that derives semantic supervision from simulator states, enabling large-scale F/T--language pretraining.
\item We characterize reuse and its limits through real-world contact recognition without encoder updates and data-efficient adaptation to an unseen insertion-completion event.
\end{itemize}
We release the code and dataset in our
\href{https://github.com/KawaharaLab/language-grounded-ft}{GitHub repository}.

\section{Related Work}
\label{sec:related_work}
\subsection{Force and Tactile Perception of Extrinsic Contact}
Extrinsic contact can be estimated and controlled using tactile and kinematic observations at the grasp, including contact location and transmitted wrench~\cite{ExtrinsicContact}. Such work shows that contact away from the fingertips can be inferred through the grasped object. Learned representations incorporating force have also supported manipulation: Lee et al.\ use self-supervised visual, force, and robot-state representations for insertion~\cite{VisionTouch}. More recently, ForceVLA, ForceVLA2 and FAVLA integrate force into action generation~\cite{ForceVLA, ForceVLA2, FAVLA}, and FM-VLA uses pretrained F/T histories as contact-related memory~\cite{FM-VLA}.

We share the interest in extrinsic contact but study F/T-only perception before contact-geometry estimation or policy learning. We evaluate whether a common representation supports language-based identification of several physical states and adaptation to an unseen contact event.

\subsection{Tactile Pretraining and Language Alignment}
Sparsh, Sparsh-X, and AnyTouch 2 learn tactile representations of physical attributes and contact dynamics through self-supervision~\cite{Sparsh, Sparsh-X, Anytouch2}. UniTouch and TVL connect touch with vision and language~\cite{unitouch, TVL}, while CLTP and FG-CLTP align tactile 3D point clouds with contact descriptions~\cite{CLTP, FG-CLTP}. CLTP and FG-CLTP are particularly related in their use of simulated contacts, attribute-based descriptions, and transfer to real sensing. OmniVTLA semantically aligns tactile information including spatial contact-force fields~\cite{OmniVTLA}; TaF-VLA uses force information to supervise an encoder of tactile histories~\cite{TaF-VLA}.

Our input is the history of resultant forces and torques at each finger, rather than contact images, point clouds, or spatial force distributions. We share the use of semantic alignment but evaluate recognition on unseen objects and transfer to real extrinsic contact.

\subsection{Sensor--Language Representations}
IMU2CLIP aligns inertial time series with video and language~\cite{IMU2CLIP}, and ImageBind connects heterogeneous modalities, including IMU signals, in a shared space~\cite{ImageBind}. SensorLM generates statistical, structural, and semantic descriptions for large-scale sensor--language pretraining~\cite{SensorLM}. For force signals, Tejwani et al.\ combine contrastive and reconstruction/translation learning to relate human-applied force profiles to language~\cite{ForceLanguage}.

Our target is the physical state arising from robot--object--environment interaction. We construct window-level descriptions from simulator states and evaluate both language-prototype recognition and supervised adaptation to a new event.

\section{Language-Grounded F/T Pretraining}
\label{sec:method}
\subsection{Input and Shared Representation}
We learn a representation of F/T histories for physical-state readout and reuse across objects and contact events. The history up to time $t$ is
\begin{equation}
\mathbf{X}_t=[\mathbf{x}_{t-T+1},\ldots,\mathbf{x}_t]
\in\mathbb{R}^{T\times D_{\mathrm{sensor}}},
\end{equation}
where $D_{\mathrm{sensor}}=12$ concatenates six-axis measurements from both fingers. The F/T encoder maps this history to $\mathbf{z}^{\mathrm{F/T}}_t=f^{\mathrm{F/T}}_\theta(\mathbf{X}_t)\in\mathbb{R}^{d}$. Sentence-BERT~\cite{SBERT} maps the paired description $l_t$ to $\mathbf{z}^{\mathrm{Lang}}_t=f^{\mathrm{Lang}}(l_t)\in\mathbb{R}^{d}$. We freeze the language encoder and optimize only the F/T encoder. 

\subsection{Contrastive Alignment and Readout}
For a batch of $B$ pairs, we define
\begin{equation}
s_{ij}=\frac{\operatorname{cos}(\mathbf{z}^{\mathrm{F/T}}_i,\mathbf{z}^{\mathrm{Lang}}_j)}{\tau},
\end{equation}
where the temperature is fixed at $\tau=0.07$, and cosine similarity uses normalized embeddings. The symmetric InfoNCE objective~\cite{CPC, CLIP} is
\begin{align}
\mathcal{L}_{\mathrm{F/T}\rightarrow\mathrm{Lang}}
&=-\frac{1}{B}\sum_{i=1}^{B}\log\frac{\exp(s_{ii})}{\sum_{j=1}^{B}\exp(s_{ij})},\\
\mathcal{L}_{\mathrm{Lang}\rightarrow\mathrm{F/T}}
&=-\frac{1}{B}\sum_{i=1}^{B}\log\frac{\exp(s_{ii})}{\sum_{j=1}^{B}\exp(s_{ji})},\\
\mathcal{L}_{\mathrm{align}}
&=\tfrac{1}{2}\bigl(\mathcal{L}_{\mathrm{F/T}\rightarrow\mathrm{Lang}}+\mathcal{L}_{\mathrm{Lang}\rightarrow\mathrm{F/T}}\bigr).
\end{align}

At inference, candidate descriptions are encoded as language prototypes, and the state with maximum cosine similarity is selected. Comparing a history with ``Holding an object.'' and ``Placing an object.'' thus provides a state readout without a trained classification head. For an event absent from pretraining, we instead attach a supervised head and evaluate frozen linear readout and LoRA adaptation. These protocols distinguish querying known semantics from learning a new event.

\section{Generating F/T--Language Pairs from Physics}
\label{sec:dataset}

\subsection{Interactions and Observations}
We automate pick-and-place with a Franka Emika Panda in Genesis~\cite{Genesis}, varying objects, density, motion conditions, and supports.
Episodes include grasping, lifting, holding, descent, placement, pressing, and release, with finger--object slip and object--support contact.

Two virtual six-axis fingertip sensors record 12-dimensional F/T at \qty{100}{\Hz}.
We extract causal \qty{0.8}{\s} histories (80 samples) at a \qty{0.1}{\s} stride.
Synchronized wrist-view and fixed third-person RGB images are stored but are not inputs to training or evaluation.

\subsection{Semantic Descriptions}
\label{subsec:semantic_annotation}
Simulator states determine four attributes, composed into descriptions according to the window's manipulation and contact conditions:
\begin{itemize}
\item \textbf{Action:} Command logs and contact events define the state. Lift contains liftoff, place contains initial support contact, and press denotes subsequent sustained pressing. Hold includes transporting an object in the air. 
\item \textbf{Weight:} Light or heavy based on mass, omitted when the environment fully supports the object.
\item \textbf{Interaction:} Stable, slip, or accidental drop, based on relative finger--object motion and contact. Stable/slip is omitted when grasp contact is insufficiently sustained.
\item \textbf{Support Surface:} Flat, step, slope, or wobbling support during placement or pressing, based on the surface actually contacted and its motion. 
\end{itemize}
Windows with missing attributes are retained if other attributes remain valid.
Descriptions include ``Lifting a heavy object while it is slipping.'' and ``Placing an object on a wobbling support.''; transitions appear as, for example, ``place then release.''
Simulator states supply supervision without inferring labels from F/T alone; we do not assume that every internal attribute is uniquely recoverable from F/T.

\subsection{Scale and Splits}
We retain accidental drops as physical failure events but exclude episodes outside common force/position quality criteria, including forces of at least \qty{100}{\N} or displacements of at least \qty{5}{\cm} within \qty{0.01}{\s}.
After filtering, the collection contains 1,531,185 windows from 40,878 episodes.
Distribution-aware subsampling without replacement reduces dominance by long-lasting states while retaining rare events and transitions.
The pretraining set contains 437,244 windows from 40,851 episodes involving 892 Google Scanned Objects (GSO)~\cite{GoogleScanObject}.

Evaluation uses 111,355 windows from 2,173 episodes involving 75 YCB objects~\cite{YCB} absent from training.
Attribute-specific subsets yield 19,683 window evaluations in total.
Encoder checkpoints are selected using a window-level validation split from GSO; the object-disjoint YCB set is used only for final evaluation.

\section{Experiments}
\label{sec:eval}

\begin{table*}[t]
    \centering
    \caption{State recognition for unseen objects. Macro F1 scores are means$\pm$sample standard deviations across 3 seeds, reported as percentages. The first three models use sentence-template text prototypes; the last is a supervised classifier trained on the same attributes, including support surface.}

    \label{tab:model_comparison}
    \footnotesize
    \begin{tabular*}{\textwidth}{@{\extracolsep{\fill}}lccc@{}}
        \toprule
        Model & Action & Weight & Interaction \\
        \midrule
        PatchTST & $94.02\pm0.11$ & $93.00\pm0.50$ & $78.14\pm0.73$ \\
        CNN                   & $93.28\pm0.19$ & $\mathbf{93.97\pm0.38}$ & $81.24\pm0.46$ \\
        MLP                   & $\mathbf{94.12\pm0.31}$ & $93.38\pm0.50$ & $\mathbf{82.57\pm0.34}$ \\
        \midrule
        Supervised MLP & $89.31\pm0.34$ & $91.12\pm0.50$ & $75.34\pm0.82$ \\
        \bottomrule
    \end{tabular*}
\end{table*}

\begin{table*}[t]
    \centering
    \caption{Linear probing after self-supervised pretraining. Macro F1 scores are reported as percentages. Results are from a single seed per method. }
    \label{tab:ssl_linear_probe}
    \footnotesize
    \begin{tabular*}{\textwidth}{@{\extracolsep{\fill}}lccc@{}}
        \toprule
        Method & Action & Weight & Interaction \\
        \midrule
        BYOL                  & $76.88$ & $83.21$ & $73.84$ \\
        Masked reconstruction & $\boldsymbol{85.24}$ & $\boldsymbol{86.81}$ & $\boldsymbol{78.30}$ \\
        VAE                   & $82.41$ & $83.71$ & $77.19$ \\
        VICReg                & $80.14$ & $73.95$ & $73.49$ \\
        \bottomrule
    \end{tabular*}
\end{table*}
We evaluate three research questions:
\begin{itemize}
\item \textbf{RQ1:} Can language prototypes read out robot action, object weight, and grasp interaction on unseen objects?
\item \textbf{RQ2:} Can the encoder recognize real contact and support conditions without parameter updates?
\item \textbf{RQ3:} Does pretraining help learn insertion completion, an event absent from its annotations, with few labeled training episodes?
\end{itemize}
RQ1 probes load and grasp information; RQ2 and RQ3 evaluate reuse for extrinsic contact.

\paragraph{Encoder implementation}
PatchTST~\cite{PatchTST} divides time series into patches and independently encodes each channel with a shared Transformer for forecasting and representation learning.
We divide each 80-sample channel into 8 patches and aggregate all 96 channel--patch features using a Token-Fusion head.
Learnable channel and patch-position embeddings are added before a one-layer Transformer with a CLS token; the CLS output is projected to 768 dimensions for language alignment. PatchTST below denotes this Token-Fusion variant.
The CNN processes the 12 input channels with four 1D convolutional blocks (kernel size 5), followed by global average pooling and a linear projection to 768 dimensions.
The MLP receives the $80\times12$ input flattened into a vector while preserving temporal and channel ordering.
All three encoders have approximately 1.11M parameters.

\paragraph{Input standardization}
We standardize each channel using training-set means and standard deviations shared across all windows.
We do not center and scale each window independently, thereby preserving between-window differences in steady loads and amplitudes.
Calibration statistics and offset correction for real data are specified below.

\subsection{RQ1: Physical-State Recognition on Unseen Objects}
\label{subsec:semantic_recognition}
\subsubsection{Protocol}
\label{subsec:classification}
We first test whether the pretrained representation preserves multiple physical attributes on objects unseen during training. Using cosine similarity to language prototypes, we recognize six actions (grasp, hold, lift, place, press, release), two weight classes (light, heavy), and three grasp-interaction classes (stable, slip, accidental drop). For action evaluation, we exclude entire episodes containing accidental drops and transition windows such as ``grasp then lift''. Evaluation sets are balanced independently for each attribute: 1,668 windows per action, 3,000 per weight class, and 1,225 per interaction class. Candidate classes are expressed using templates such as ``\{Action\} an object.'' and ``The object is \{Weight\}'', and their embeddings are compared with the F/T embedding using cosine similarity.

We compare PatchTST, CNN, and MLP under the same contrastive conditions. As a non-language semantic-supervision reference, we train a multi-task MLP on simulator-derived action, weight, grasp-interaction, and support-surface labels using the same training data.
Models use the same training data, batch size 512, learning rate $2.0\times10^{-3}$, and at most 400 epochs, with warmup, cosine decay, early stopping, and validation-based checkpoint selection. Input normalization uses training data only.

\subsubsection{Language-Prototype Readout}
All three contrastively trained encoders achieve strong Macro F1 scores on unseen objects (Table~\ref{tab:model_comparison}), indicating that the language-aligned representation preserves action, weight, and grasp-interaction information beyond the training objects.
With the same MLP backbone, language-prototype readout exceeds the supervised reference by 4.81, 2.26, and 7.23 percentage points in action, weight, and interaction Macro F1, respectively, even when the supervised targets include support surface.
One possible explanation is that fixed sentence embeddings provide a shared semantic structure for joint descriptions of physical attributes, acting as an inductive bias for generalization to unseen objects. 
Thus, in this setting, language-based readout does not require sacrificing recognition performance relative to dedicated classification heads.

No architecture is consistently best: MLP performs best on action and interaction, whereas CNN is best on weight.
These results indicate that the observed recognition capability is not restricted to a particular encoder architecture. The evaluation uses finite canonical descriptions and does not test generalization to unrestricted language queries.

\subsubsection{Self-Supervised Linear Readout}
\label{subsec:ssl_probe}
We next ask how much physical-state information can be learned from the F/T signal alone, without semantic supervision. Using the same PatchTST backbone, we compare masked reconstruction, BYOL~\cite{BYOL}, VAE~\cite{VAE}, and VICReg~\cite{VICReg}, and evaluate each frozen encoder with a supervised linear head.

Masked reconstruction predicts masked samples; BYOL matches two augmented views; VAE learns a reconstruction-based latent variable model; and VICReg regularizes invariance, variance, and covariance across augmented views.

\begin{figure*}[t]
    \centering
    \includegraphics[width=\linewidth]{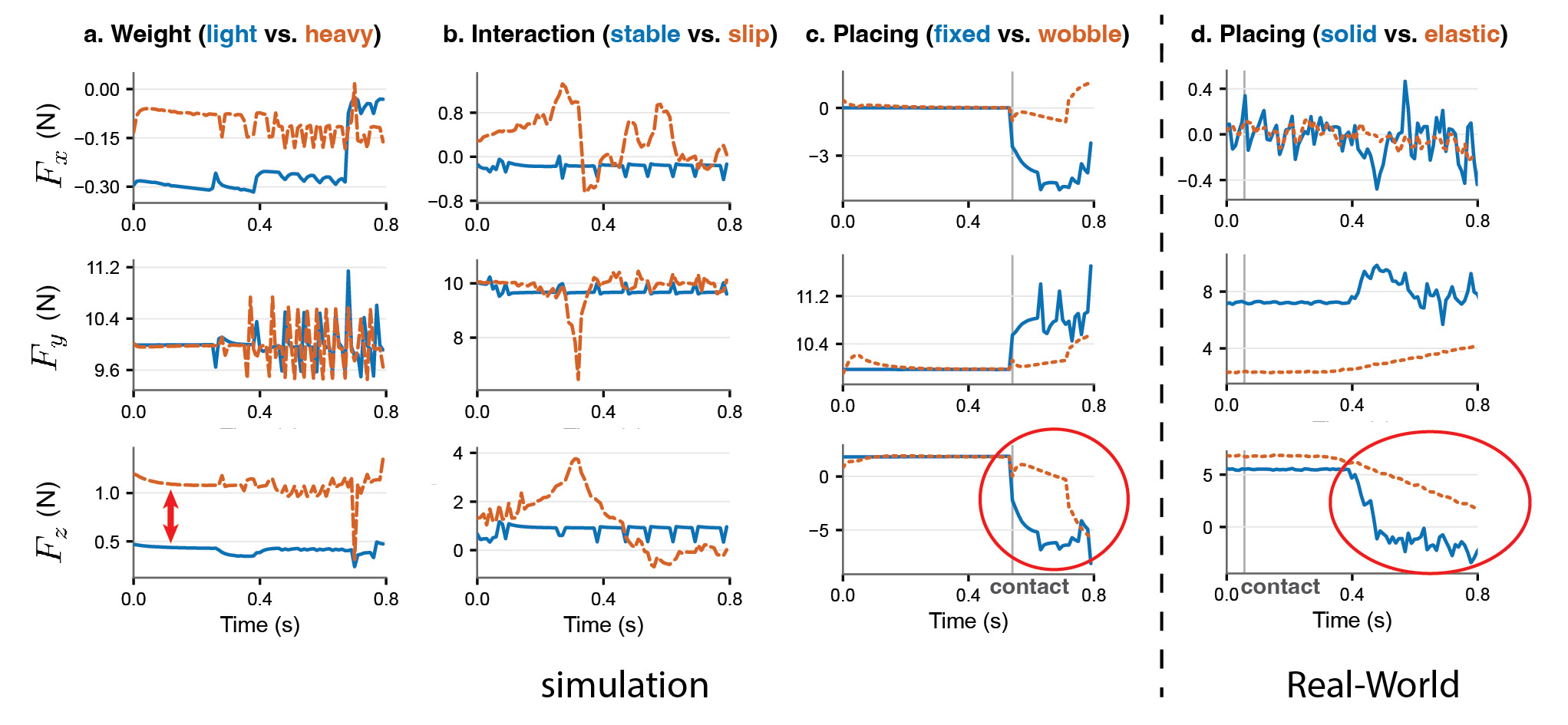}
    \caption{Example \qty{0.8}{\s} force histories under contrasting physical conditions in simulation and real-world experiments: (a) larger force magnitudes for a heavier object, (b) stronger force fluctuations during slip, (c) slower load transfer on a wobbling support, and (d) slower load transfer for a compliant object (ball).}
    \label{fig:slip_stable_ft}
\end{figure*}

Masked reconstruction gives the strongest linear readout across all three attributes, with VAE next (Table~\ref{tab:ssl_linear_probe}). This shows that signal-centric pretraining alone can recover substantial physical-state structure from F/T histories and provides a strong baseline for semantic pretraining.
Figure~\ref{fig:slip_stable_ft} qualitatively illustrates that different
physical states can produce distinct F/T dynamics.

However, Tables~\ref{tab:model_comparison} and~\ref{tab:ssl_linear_probe} differ in both readout and supervision. Their numerical differences alone do not establish general superiority of language-contrastive pretraining. RQ3 compares these objectives under common downstream adaptation protocols.

\begin{figure}[t]
    \centering
    \includegraphics[width=\linewidth]{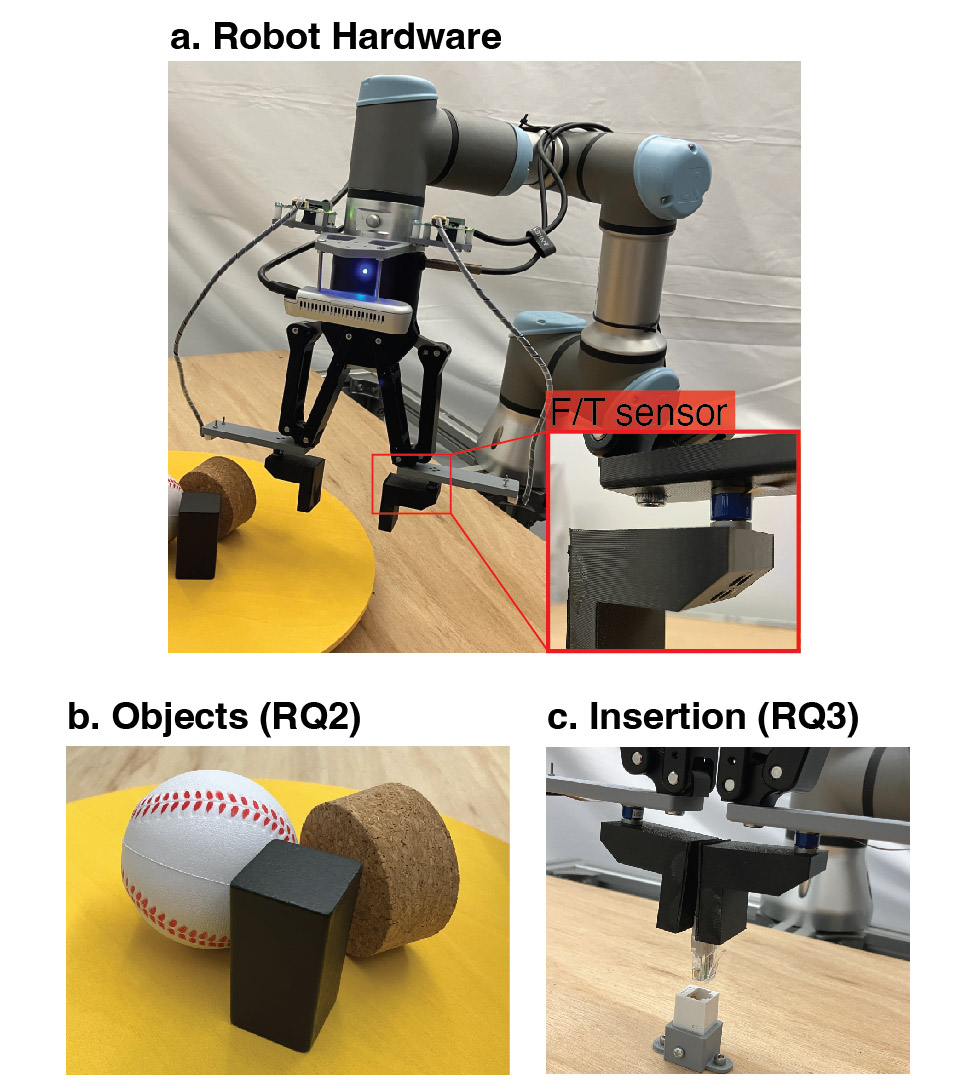}
    \caption{Hardware used in RQ2, RQ3. (a) UR3 with a Robotiq 2F gripper and bilateral MMS101 six-axis fingertip sensors, (b) Objects used in RQ2, (c) Connector and socket used in RQ3.}
    \label{fig:hardware}
\end{figure}

\subsection{RQ2: Real Extrinsic Contact without Encoder Updates}
\label{subsec:real_transfer}

We next test whether physical states learned in simulation can be read out from real F/T measurements without updating the encoder.
We evaluate two aspects of extrinsic contact: detecting support contact during placement and identifying the contacted support condition.
We use the language-pretrained PatchTST as the real-robot backbone, based on an auxiliary hold-versus-place prototype readout.
CNN and MLP were evaluated afterwards as an additional architecture check on the fixed-support wooden-cuboid task, achieving 11/20 and 17/20 detections, respectively, compared with 20/20 for PatchTST. These results were consistent with the choice of PatchTST for the main real-robot evaluation.

\subsubsection{Calibration}
To account for the simulation-to-real difference in sensor coordinates, scale, and offset, we perform calibration using unlabeled real data.
A UR3 with a Robotiq 2F gripper and bilateral MMS101 six-axis fingertip sensors (Fig.~\ref{fig:hardware}a) records 12-dimensional F/T at 100~Hz.
Sensor coordinates are mapped to the simulated fingertip frames.
Ten unlabeled real pick-and-place trials provide normalization statistics. To compensate for sensor drift, we correct offsets using \qty{1}{\s} of unloaded measurements before each grasp.
Encoder parameters remain fixed.

\subsubsection{Contact Detection}
During descent, the preceding \qty{0.8}{\s} of F/T history is processed every \qty{50}{\ms} and compared with the text prototypes ``Holding an object.'' and ``Placing an object.''
After an initial \qty{2}{\s} gate, descent is stopped when the similarity to ``Placing an object.'' exceeds that of ``Holding an object.'' for three consecutive evaluations.

As a tuning-free rule-based reference, we use the vertical-force transition caused by load transfer after support contact. 
The detector stops when the sum of the bilateral offset-corrected $F_z$ signals remains below \qty{0}{\N} for three consecutive evaluations.
This corresponds to the transition from supporting the object's weight to receiving a reaction force from the support.

Reference contact time is estimated from the operator's visual timestamp after subtracting a separately measured \qty{300}{\ms} reaction time.
We evaluate 20 trials per condition: three objects on a fixed support (wooden cuboid, cork cylinder, and compliant ball) and the wooden cuboid on a wobbling support (Fig.~\ref{fig:hardware}b).
Success requires issuing a halt command within \qty{0.8}{\s} after reference contact, matching the input-history length; beyond this interval, the initial contact transition no longer appears in the causal input window.
Halt command latency is reported over successful trials.

\begin{table}[t]
    \centering
    \caption{Real-world contact detection. Delay is measured from reaction-time-corrected contact to issuing a halt command and averaged over successful trials only.}
    \label{tab:real_contact_detection}
    \small
    \begin{tabularx}{\linewidth}{@{}Xlrr@{}}
        \toprule
        Condition & Method & Success & Delay [s] \\
        \midrule
        Fixed, wooden cuboid & Proposed & 20/20 & 0.348 \\
                             & $F_z$    & 20/20 & 0.492 \\
        \addlinespace
        Fixed, cork cylinder & Proposed & 19/20 & 0.430 \\
                             & $F_z$    & 20/20 & 0.547 \\
        \addlinespace
        Fixed, compliant ball & Proposed & 11/20 & 0.631 \\
                              & $F_z$    & 0/20  & -- \\
        \addlinespace
        Wobbling, wooden cuboid & Proposed & 12/20 & 0.437 \\
                                & $F_z$    & 0/20  & -- \\
        \bottomrule
    \end{tabularx}
\end{table}

Across the three fixed-support objects, the proposed method succeeds in 50/60 trials (83.3\%), compared with 40/60 (66.7\%) for the $F_z$ rule (Table~\ref{tab:real_contact_detection}).
The two methods are comparable on the rigid wooden and cork objects, but their behavior diverges on the compliant ball and wobbling support: the $F_z$ rule never triggers within \qty{0.8}{\s}, whereas the proposed method succeeds in 11/20 and 12/20 trials, respectively. On the wooden and cork objects, the proposed method also yields shorter mean halt-command delays over successful trials.

The failure of the $F_z$ rule under compliant and moving-support conditions is consistent with delayed load reversal after initial contact. On the wobbling support, extending its observation horizon to \qty{5}{\s} yields 15/20 eventual detections, but with a mean delay of \qty{3.890}{\s}.
The pretrained model may therefore use F/T changes during load transfer before sign reversal occurs.

\subsubsection{Support-Condition Readout}
We next test whether the same embedding used for contact detection also retains information about the contacted support. We compare the prototypes ``Placing an object on a fixed flat surface.'' and ``Placing an object on a wobbling surface.'' on successfully detected wooden-cuboid contacts: 20 fixed-support and 20 wobbling-support trials. 
The latter comprise the initial 12 successes plus 8 successes obtained from 13 additional attempts.
The support is predicted from whichever prototype has higher cosine similarity, with no trained classification head or tuned threshold.

\begin{table}[t]
    \centering
    \caption{Support recognition conditional on successful contact detection. Similarities are mean cosine similarities to the two language prototypes.}
    \label{tab:real_support_classification}
    \small
    \begin{tabularx}{\linewidth}{@{}Xrrr@{}}
        \toprule
        True support & fixed cosine & wobble cosine & Accuracy \\
        \midrule
        Fixed & 0.263 & 0.145 & 18/20 (90.0\%) \\
        Wobbling & 0.061 & 0.081 & 13/20 (65.0\%) \\
        \midrule
        Overall & -- & -- & 31/40 (77.5\%) \\
        \bottomrule
    \end{tabularx}
\end{table}

The prototype comparison correctly identifies 31/40 trials (77.5\%, Table~\ref{tab:real_support_classification}).
Thus, the contact-detection embedding retains information about support condition. This result is conditional on successful contact detection, and wobbling supports remain less reliable (13/20).

\subsection{RQ3: Adapting to Unseen Insertion Completion}
\label{subsec:fewshot_insertion}
We finally test whether pretraining helps learn a new extrinsic-contact event that is not represented in the pretraining annotations. The downstream task is to recognize when a connector becomes fully seated, which can inform whether insertion should continue or stop (Fig.~\ref{fig:hardware}c).

\subsubsection{Data and Protocol}
We formulate insertion completion as binary recognition of whether the connector is fully seated, rather than insertion-motion generation. Using separately collected real-robot data, we record bilateral F/T at \qty{100}{\Hz} and extract causal \qty{0.8}{\s} histories every \qty{0.1}{\s}.
The reference seating time $t_{\mathrm{seat}}$ is obtained by subtracting a measured \qty{300}{\ms} reaction time from the operator timestamp, and each window is labeled by its endpoint:
\begin{equation}
 y_t=\begin{cases}
 0, & t<t_{\mathrm{seat}},\\
 1, & t\geq t_{\mathrm{seat}}.
 \end{cases}
\end{equation}
Thus, the positive class includes full seating and all subsequent observations.

We split 125 successful episodes into 100 training, 5 validation, and 20 held-out test episodes.
We vary the number of labeled training episodes $N\in\{1,2,5,10,25,50,100\}$ over five seeds, using identical nested subsets across methods.
Normalization statistics are fixed from the full 100-episode training split; checkpoints are selected on the validation set, and evaluation uses window-level Macro F1 on the test episodes.

\subsubsection{Adaptation Methods}
Using the same PatchTST backbone, we compare language-contrastive, masked-reconstruction, VAE, VICReg, and BYOL pretraining. Frozen readout trains only a 769-parameter linear head.
LoRA~\cite{LoRA} updates rank-4 adapters in the attention query/value projections together with the head, for 8,961 trainable parameters. Scratch End-to-End instead trains all 1,111,937 parameters from random initialization on the downstream data.

\begin{figure}[t]
    \centering
    \includegraphics[width=\linewidth]{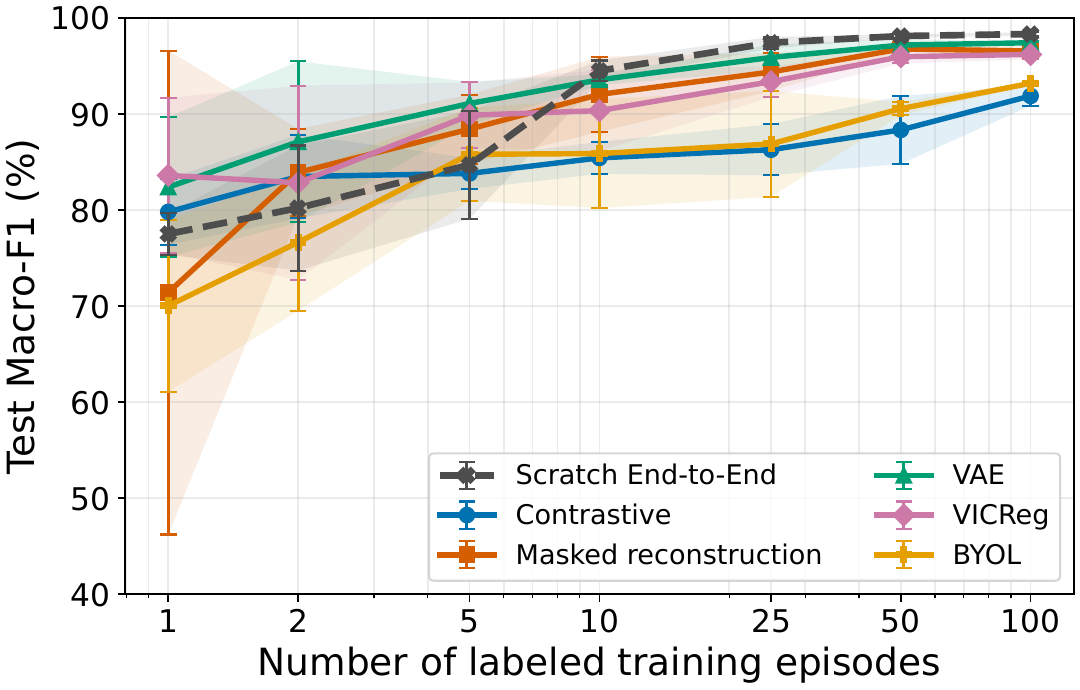}
    \caption{Insertion-completion recognition with frozen linear readout and Scratch End-to-End. Curves show five-seed means; bands and error bars indicate $\pm1$ standard deviation. }
    \label{fig:terminal_insertion_linear_vs_scratch}
\end{figure}

\subsubsection{Limits of Frozen Readout}
Frozen pretrained representations help only in the very low-data regime
(Fig.~\ref{fig:terminal_insertion_linear_vs_scratch}).
For example, at $N=1$, VICReg and VAE reach $(83.62\pm8.08)\%$ and $(82.40\pm7.27)\%$, compared with
$(77.52\pm2.15)\%$ for Scratch.
By $N=10$, however, Scratch reaches $(94.51\pm1.08)\%$ and exceeds the best frozen representation, and this gap persists at larger $N$. At $N=100$, Scratch reaches $(98.34\pm0.19)\%$, compared with $(91.90\pm1.02)\%$ for frozen language-contrastive features.

Thus, representations that support known semantic readout do not necessarily make a new event linearly separable. This motivates adapting the representation rather than keeping it fixed.

\begin{figure}[t]
    \centering
    \includegraphics[width=\linewidth]{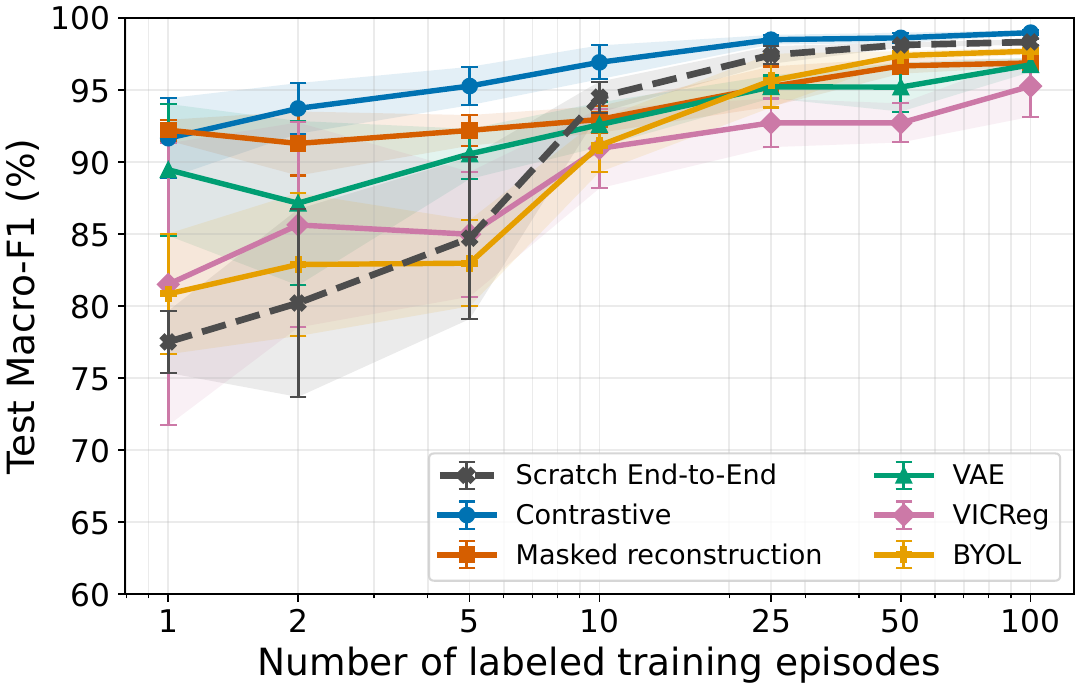}
    \caption{Insertion-completion recognition with LoRA adaptation. Evaluation conditions and the Scratch baseline are shared with Fig.~\ref{fig:terminal_insertion_linear_vs_scratch}. Curves show five-seed means; error bars and bands indicate $\pm1$ sample standard deviation. 
    }
    \label{fig:terminal_insertion_lora_vs_scratch}
\end{figure}

\subsubsection{Parameter-Efficient Adaptation}
Adapting the pretrained representation changes this trend (Fig.~\ref{fig:terminal_insertion_lora_vs_scratch}).
With LoRA, language-contrastive pretraining reaches $(91.66\pm2.77)\%$ at $N=1$, $(95.29\pm1.33)\%$ at $N=5$, and $(96.93\pm1.17)\%$ at $N=10$, compared with Scratch means of $(77.52\pm2.15)\%$, $(84.72\pm5.61)\%$, and $(94.51\pm1.08)\%$, respectively.
The advantage narrows as more labeled data become available, with both models approaching 99\% at $N=100$.

The benefit is not unique to language supervision: masked reconstruction has the highest mean at $N=1$, and other pretrained objectives also benefit from adaptation. These results indicate that pretraining is more useful for this new event as an adaptable initialization than as a fixed feature extractor.
Together, RQ2 and RQ3 characterize two forms of reuse: querying known semantics under new conditions and adapting a pretrained representation to a new contact event.

\section{Limitations and Future Work}
\label{sec:limitations}
The scope of pretraining remains limited. Our simulated data center on pick-and-place and do not cover the full diversity of contact-rich manipulation, including sustained sliding, rotation,
deformation, and broader variations in object and support properties. Real-robot deployment also requires calibration from unlabeled data, and contact-detection failures remain for compliant objects and moving supports. Future work should combine broader simulation with automatically collected real interactions to reduce this domain gap.

The evaluation also limits the conclusions that can be drawn. 
Language-based recognition uses a finite set of predefined prototypes and does not test paraphrase robustness or unseen concepts.
Support-condition recognition is evaluated only after successful contact detection, and the insertion dataset contains successful episodes only; failure discrimination and recovery are therefore not established. 

\section{Conclusion}
\label{sec:conclusion}
We studied language-grounded pretraining of bilateral F/T histories for reusable physical-state perception. Simulator states generate supervision for 437,244 paired windows from 40,851 episodes and 892 objects.
Language prototypes read out multiple attributes on unseen objects and recognize real contact and support conditions after unlabeled calibration, without encoder updates. 
For unseen insertion completion, frozen features have limitations, but LoRA improves performance with few labeled training episodes. These findings show that low-dimensional F/T histories transmitted through a grasped object can support both known-state readout and adaptation to new extrinsic contact events.

\appendix
\section{Reproducibility Notes}
\label{app:repro}

\paragraph{Wrench and annotation conventions}
For each simulated finger, the recorded wrench sums all contact forces on the finger link in its local frame, with torque computed about the link origin. Real measurements use the same channel order and an axis mapping $R_z(\pi)$, but no torque-origin translation; thus, exact geometric equivalence of simulated and physical torque reference points is not assumed. Contact-state transitions are confirmed over five consecutive samples and assigned to the first sample of the confirmed run. Weight is split at \qty{0.2}{\kg}; slip is defined from tangential finger--object relative motion with a \qty{5}{\mm/s} threshold.

\paragraph{Language training and splits}
The retained corpus contains a finite template vocabulary and uses the frozen \texttt{all-mpnet-base-v2} Sentence-Transformer.
InfoNCE treats only the paired diagonal example as positive; windows with identical descriptions may remain in the denominator.
Historical checkpoints use a window-level GSO validation split. The final YCB evaluation uses disjoint objects and is not used for checkpoint selection.
In an additional three-seed MLP sensitivity study, episode-disjoint validation yielded similar YCB Macro F1 scores to the historical window-level split (Action: $-0.20$ pp, Weight: $-0.03$ pp, Interaction: $+0.59$ pp).

\paragraph{Computational resources and training time}
Training was ran on a workstation with an NVIDIA RTX A6000 GPU. 400-epoch contrastive training with batch size 512 took 7.71--7.89 h for PatchTST, 0.98--1.03 h for CNN, and 0.66--0.70 h for MLP.

\paragraph{Inference latency}
An additional offline benchmark of the deployed PatchTST readout on one RTX A6000 (batch size one, $80\times12$ input) measured 1.83 ms mean and 1.89 ms 95th-percentile latency over 1,000 calls after 200 warm-up calls.
In the online inference, the interval averaged 11.18 ms (95th percentile: 15.61 ms) over all placement trials.

\section*{Acknowledgment}
This work was supported by JSPS KAKENHI Grant Number JP25H01135. We gratefully acknowledge this support.

\bibliographystyle{IEEEtran}
\bibliography{ref}
\end{document}